\documentclass{article}

\usepackage{url}
\usepackage{setspace}
\usepackage[top=3cm, bottom=3cm, left=3cm, right=3cm]{geometry}

\usepackage{graphicx}

\title{CoMOGrad and PHOG: From Computer Vision to Fast and Accurate Protein Tertiary Structure Retrieval}

\author{Rezaul Karim $^{1}$ , Mohd. Momin Al Aziz$^{1}$ , Swakkhar Shatabda $^{2}$ ,
M. Sohel Rahman $^{1}$,\\ Md. Abul Kashem Mia$^{1}$ and Farhana Zaman$^{1}$, Salman Rakin$^{1}$\\$^{1}$Department of Computer Science and Engineering,\\
 Bangladesh University of Engineering and Technology,\\ Palashi, Dhaka-1000, Bangladesh.\\
$^{2}$Department of Computer Science and Engineering,\\ United International University,\\ House 80, Road 8A, Dhanmondi, Dhaka-1209, Bangladesh.}
\date{}
\newcommand{\concern}[1]{{#1}}
\newcommand{\change}[2]{{#2}}
\newcommand{\add}[1]{{#1}}
\newcommand{\remove}[1]{{}}

\begin{document}

\maketitle
\doublespacing

\begin{abstract}
Due to the advancements in technology number of entries in the structural database of proteins are increasing day by day. Methods for retrieving protein tertiary structures from this large database is the key to comparative analysis of structures which plays an important role to understand proteins and their function. In this paper, we present fast and accurate methods for the retrieval of proteins from a large database with tertiary structures similar to a query protein. Our proposed methods borrow ideas from the field of computer vision. The speed and accuracy of our methods comes from the two newly introduced features, the co-occurrence matrix of the oriented gradient and pyramid histogram of oriented gradient and from the use of Euclidean distance as the distance measure. Experimental results clearly indicate the superiority of our approach in both running time and accuracy. Our method is readily available for use from this website: \url{http://research.buet.ac.bd:8080/Comograd/}.
\end{abstract}

\section{\label{intro}Introduction}

Proteins perform a large array of functions within a living organism. Proteins are polymers of amino-acid monomers. Linear sequence of amino acids is called the primary structure of a protein. There are three other levels of structural complexity: secondary, tertiary and quaternary. Secondary structure is the local spatial arrangement of the backbone atoms  formed by intramolecular and intermolecular hydrogen bonding of amide groups.  Tertiary structure  refers to the three-dimensional structure of an entire polypeptide and quaternary structure  is the spatial arrangement of two or more polypeptide chains known as sub-units.

Every protein has a unique, stable and kinetically accessible \cite{anfinsen} three dimensional structure or tertiary structure also known as \add{the} native structure. The functionalities of a protein are principally correlated to its tertiary structure. That is why a protein ceases to function when its tertiary structure is broken by denaturation at a high temperature; although, its primary structure may still remain intact \cite{tanford1968protein}. Proteins with similar tertiary structure have similar ligand binding sites and pockets \cite{wass2009prediction}. Moreover, tertiary structures are more conserved than amino acid sequence during evolution \cite{illergaard2009structure}. Misfolded proteins are the cause of many critical diseases like Alzheimer's disease \cite{detoma2012misfolded}. Therefore, analysis and study of similar tertiary structures is of great importance in function prediction of novel protein\add{s}, study of evolution, disease diagnosis, drug discovery, antibody design and many other fields.

Following the determination of the first tertiary structure in 1958 using crystallography \cite{kendrew1958three}, biologists have successfully determined a large number of structures by now. Due to recent developments in X-ray crystallography and nuclear magnetic resonance imaging, there has been a rapid increase in the number of experimentally determined structures stored in the world-wide repository\add{,} Protein Data Bank (PDB)~\cite{PDB} (\url{http://wwpdb.org/}). RCSB PDB is the primary repository for known protein structures. As of February 11, 2014, the number of protein structures stored in PDB was more than 97,789. With this increase in the number of known protein structure\add{s}, the need \change{of}{for} fast and accurate algorithms for the retrieval of protein tertiary structures is now greater than ever. In the protein tertiary structure retrieval problem, the task is to retrieve proteins having most similar structures from the known structure database. The results are ranked based on similarity or distance measures. 

There exist numerous approaches in the literature that focus on fast and accurate retrieval of protein tertiary structures. The traditional way to compare structures is to treat each one as a rigid three-dimensional object and superimpose one on the other. Differences are then calculated using different distance metrics, e.g., least-squares method. In the pattern recognition literature, the structures are often represented by feature vectors and similarity or dissimilarity is measured by comparing the feature vectors with one another. As a feature to represent the tertiary structure of a protein chain, the $\alpha$ carbon distance matrix is widely used by many researchers \cite{singh1997hierarchical,holm1997dali}. Here, the 3D coordinate data is mapped to a two dimensional feature matrix. The $\alpha$ carbon distance matrix gives the intra-molecular distances of $\alpha$ carbons in a protein chain. This $\alpha$ carbon distance matrix resembles the tertiary structure of a protein and the secondary structure elements conserved in it. Notably, $\alpha$ carbon distance matrix based exact algorithms run in $O(N!)$ time assuming that the input matrix is of size $N\times N$.

$DALI$~\cite{holm1997dali} compares structures by aligning the $\alpha$ carbon distance matrices. Each distance matrix is decomposed into sub-matrices of fixed size which is called the elementary contact patterns. It compares those contact patterns (pair-wise), and store the matching pairs in a list with a matching score. Then, it assembles pairs in the correct order using Monte Carlo optimization to yield the overall alignment and the final matching score. In a later version, it \add{has} used branch and bound method to assemble \add{the} pairs \cite{holm1996mapping}. The $CE$ method~\cite{Shindyalov01091998} also takes the $\alpha$ carbon distance matrix as the feature vector and uses combinatorial extension and Monte Carlo optimization to compare protein structures in a way much similar to $DALI$. Both $DALI$ and $CE$ require lots of computation and hence the corresponding web servers respond to the query requests via email only after a certain period that is required for costly processing. A faster approach based on the $\alpha$ carbon distance matrix as a feature is $MatAlign$~\cite{aung2006matalign}. It provides an $O(N^4)$ dynamic programming solution where $N$ is the dimension of the distance matrix. $MatAlign$ compares distance matrix of the query protein and the target protein row by row and builds up a dynamic programming table based on a row by row matching score. Then, it aligns rows of the query protein with the rows of the target protein to maximize the corresponding matching score. The higher is the matching score, the more is the similarity in the corresponding structures. 

Marsolo et al.~\cite{marsolo2006structure} \add{have} introduced a wavelet based approach that resizes the distance matrices of the protein structures before the actual comparison is done. Later on Mirceva et al. \add{have} introduced \textit{MASASW} \cite{6051424} that uses wavelet coefficients of distance matrices as the feature vector. It has been \remove{claimed and} shown in \cite{6051424} that \textit{Daubechies-2 wavelets} improves accuracy than others. \textit{MASASW} transforms all the $\alpha$ carbon distance matrices \change{to}{into} a $32\times32$ matrix by interpolation and wavelet transformation. It then compares them like $MatAlign$ but with a sliding window to reduce the number of comparisons. The time complexity of \textit{MASASW} is $O(wWN^2)$ where $w$ and $W$ are window sizes and $N$ is the dimension of the distance matrices. As has been mentioned above\add{,} \textit{MASASW} assumes $N=32$.

Despite \add{that} several methods are found in the literature for protein structure retrieval, the quest for even faster and more accurate methods still continues as the number of known protein structures is growing very fast. In this paper\add{,} we present an extremely fast and highly accurate novel method of retrieving proteins with similar tertiary structures from a large database. In particular, here we present an ultra fast algorithm based on two novel feature vectors. These are the \textbf{Co}-occurrence \textbf{M}atrix of the \textbf{O}riented \textbf{Gra}dient of \textbf{D}istance Matrices (CoMOGrad) and \textbf{P}yramid \textbf{H}istogram of \textbf{O}riented \textbf{G}radient (PHOG). Additionally, as will be reported later, our proposed algorithm gives more accurate results than the state of the art methods. Very briefly, much of the speed and accuracy we have achieved comes from the introduction of the novel features from the field of computer vision and pattern recognition. Our aim has been to introduce a feature which does not require any complex algorithm to compare the tertiary structures. Rather, a simple distance measure to calculate \add{the} distance between \add{the} two vector quantities is used in our approach. \add{As has already been mentioned above,} \change{Some}{some} previous methods \add{in the literature have} used \add{the} $\alpha$ carbon distance matrix as \add{a} feature vector. Upon analyzing the tertiary structure and $\alpha$ carbon distance matrix represented as a gray-scale image, we have observed that not all data in the matrix seem to be equally important. We further have realized that the co-occurrence matrix (as will be reported later) of the oriented gradient of the distance matrix is the most important feature with respect to the comparison of tertiary structures. Finally, we have found that the Euclidean distance or $\ell^2$ norm of our novel features as the distance measure outperforms the widely used costly alignment distance/similarity measure of $\alpha$ carbon distance matrices. As a result, the combination of the above ideas gives us an extremely fast method without sacrificing the accuracy.

Later in this paper, we first describe our proposed approach along with the novel features. Following this, we report the experimental results and relevant discussion. Finally, we conclude the paper with an outline of the future works.





\begin{figure}
 \centering
 \includegraphics[scale=1]{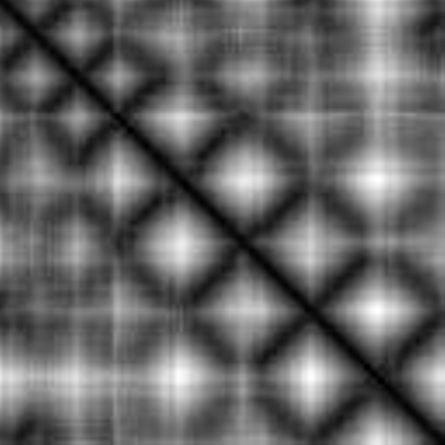}
 \caption{ Representation of $\beta$ sheets of domain d1n4ja \cite{le2003structural} in $\alpha$ carbon distance matrix gray-scale image.\label{fig:beta sheet example}}

 \end{figure}

\begin{figure}
\centering
 \includegraphics[scale=1]{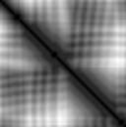}
\caption{Representation of $\alpha$ helices of domain d1irqa \cite{Murayama2001789} in $\alpha$ carbon distance matrix gray-scale image.\label{fig:alpha helix example}}
\end{figure}

We have carefully analyzed the gray-scale images from the $\alpha$ carbon distance matrices and the tertiary structures. We have observed that the $\alpha$ helices and the anti-parallel beta sheets appear as dark lines parallel to the diagonal dark line and parallel beta sheets appear as dark lines normal to the diagonal dark line. Beta sheets of two strips appear as one dark line normal to the diagonal; beta sheets of three strips appear as two dark lines normal to the diagonal and one dark line parallel to the diagonal. In general, for a standard beta sheet, the number of points of co-occurrence of parallel and anti-parallel diagonal lines depends on the number of strips in the beta sheets. \concern{Again, the number of single parallel lines depends on the number of standard $\alpha$ helices.} Moreover, length of those lines near the diagonal region is proportional to the length of the $\alpha$ helices. The distance of the parallel lines from the diagonal dark line is proportional to the radius of the $\alpha$ helix. Figure~\ref{fig:beta sheet example} depicts the corresponding $\alpha$ carbon distance matrix as a gray scale image of a tertiary structure of a protein with beta sheets. In the gray scale image, the 7 anti-parallel dark lines near the diagonal dark line correspond to the presence of 8 beta sheets in the corresponding protein structure and the lengths of those dark lines are proportional to the lengths of the beta sheets. Figure~\ref{fig:alpha helix example} represents the gray scale image of the corresponding $\alpha$ carbon distance matrix of a protein tertiary structure with two alpha helices. \add{Here in} \remove{In} the image\add{,} the two parallel dark lines near the diagonal dark line correspond to the presence of two $\alpha$ \change{helix}{helices} in the protein structure and the lengths of the dark lines are proportional to the lengths of the $\alpha$ helices.

 In the contemporary literature of computer vision and digital image processing \cite{Gonzalez:2001:DIP:559707} the lines in digital images are usually recognized from \add{the} gradient of the images. This leads us to believe that, the co-occurrence of gradient angles represents the secondary structure elements more precisely than just the distance matrix image. The tertiary structure involves the presence of secondary structure elements (SSEs), their size and position in the chain and their orientation. From the images of protein structures and their corresponding $\alpha$ carbon distance matrix gray-scale image, it is clear that the SSEs are represented by the orientation of dark lines \change{in}{at} the near diagonal region of the $\alpha$ carbon distance matrix image. The position of the SSEs in a protein chain is represented by the position of the dark lines \change{in}{at} the near diagonal region of the image. Here, \emph{near diagonal region} is the region nearby the diagonal dark line. The size of the SSEs are represented by the length of the dark line \change{in}{at} the near diagonal region. The orientation of the SSEs are represented by the presence of the dark lines \change{in}{at} the far diagonal regions and \add{the} darkness and orientation of those dark lines. Here, \emph{far diagonal region} is the region in the image that is distant from the main diagonal dark line. Therefore, we need to incorporate the gradient orientation angle along with the gradient magnitude and gradient spatial orientation to incorporate the orientation of  the SSEs in the feature vector. We also need co-occurrence of gradient angles. \add{Based on our study} \change{We}{we} introduce \textbf{Co}-occurrence \textbf{M}atrix of the \textbf{O}riented \textbf{Grad}ient (CoMOGrad) as our first feature vector which incorporates gradient orientation angles and co-occurrence of \add{the} gradient orientation angles. This feature enables implementation of an algorithm that can facilitate more than 100 times faster comparison than \textit{MASASW}. However, \change{the}{this} speed is achieved at the cost of a slight decrease in the accuracy. \change{We}{Subsequently, we} introduce another feature vector, namely, \add{the} \textbf{P}yramid \textbf{H}istogram of \textbf{O}riented \textbf{G}radient (PHOG) by incorporating \add{the} spatial information and gradient magnitude. The combination of \add{both} CoMOGrad and PHOG features in comparing structures results in an algorithm that is not only more accurate but also more than 40 times faster than \textit{MASASW} in comparing two structures. The details of the experiments are presented later in the results section.

\section{Methods and Materials}

\subsection{Feature vectors and feature extraction.}
\label{featureVector}
\paragraph{Mapping 3D coordinates to 2D function:}
\label{distmatrix}
Distance matrix of $\alpha$ carbons in residues is a good candidate to transform the 3D structure to the corresponding 2D vector representation as shown in $MASASW$~\cite{6051424} and the wavelet based approach by Marsolo et al.~\cite{marsolo2006structure}. This distance matrix \change{is}{gives} the pairwise distance between all pair\add{s} of $\alpha$ carbons in the polypeptide chain. Proteins with similar tertiary structures will have similar distance matrices and vice versa. As stated earlier, if we consider the matrix as a monochromatic image, $\alpha$-helices and parallel $\beta$-sheets will appear as dark lines parallel to the main diagonal and antiparallel beta sheets will appear as dark lines normal to \add{the} main diagonal. This distance matrix feature also has a very appealing property, i.e., this is translation, scaling and rotation invariant.

Interestingly, as it is a two dimensional matrix like a digital image we can easily apply image processing and computer vision algorithms on it. Similarly, we can use this matrix as an adjacency matrix and interpret it as a graph. Subsequently, graph theory techniques may also be applied to solve the tertiary structure retrieval problem with this feature. In this paper, we apply ideas from the field of image processing and computer vision. Most recently these ideas have got their niche in pedestrians and car detection \cite{watanabe2009co,ren2010fast}.

\subsubsection*{Scaling C$\alpha$-C$\alpha$ distance matrix images to the same dimension}
\paragraph{Bi-cubic interpolation:}
As different protein chains have different number of $\alpha$ carbons, \add{the dimensions of} their $\alpha$ carbon distance \change{matrix dimension varies}{matrices vary}. Therefore, we need to scale the distance matrices to the same dimension. For scaling the distance matrices, we use the methods of digital image processing used for image resizing. At first, we scale all the images to the dimension that is a power of 2 and nearest to their original dimension. As an example, if the image dimension is $80\times 80$ we scale it to $64\times 64$ and if the original image dimension is $100\times 100$ we scale to $128\times 128$. For this step, we use bi-cubic interpolation.

\paragraph{Wavelet transform:}
\label{wavelet_xform}
After scaling the images as mentioned above, we apply wavelet transform to transform all the images to the same dimension. Notably, wavelet transform is the most widely used technique for image scaling or resizing in digital image processing. Using wavelet transform, we scale all images to $128\times 128$ dimension as most of the images in the previous step were found to be of that dimension. For wavelet filter we used Daubechies-2 wavelet \cite{daubechies1988orthonormal} as this filter has been shown to have outperformed other traditional wavelets for protein structure feature representation by $MASASW$~\cite{6051424}. Wavelet transform of an image gives four images, namely, \add{the} \emph{approximate detail, horizontal detail, diagonal detail} and \emph{vertical detail}.  Each of these images \change{are with}{has} dimension \add{that is} half of the original image. We take the approximate detail \change{for scaling}{to scale a} large\remove{r} image\remove{s} to \add{a} smaller size since this is the approximate sub sampled image. For the images with dimension greater than $128\times 128$, we perform wavelet transform on each images multiple times to get the approximate coefficient of size $128\times 128$. For images with dimension less than $128\times 128$, we first perform wavelet transform on the images. Then using bi-cubic interpolation, we scale all four coefficients to twice their initial size. After that, with inverse wavelet transform on the four scaled wavelet coefficients, we get original image scaled up to twice its initial size. With repeated application of scaling up \add{(down)} the images that are smaller \add{(higher)} than dimension $128\times 128$\add{,} \remove{and scaling down which are greater than dimension $128\times 128$,} we finally \change{have}{get} all of them in the desired dimension of $128\times 128$.

\subsubsection*{Novel features from scaled C$\alpha$-C$\alpha$ distance matrix images}
\paragraph{Co-occurrence matrix of oriented gradient (CoMOGrad):}
\label{co-occurrence matrix of oriented gradient of wavelet coefficients}
After having all the images of dimension $128\times 128$ we extract our CoMOGrad feature. First we take \add{the} gradient of each of the images and compute the gradient angle and magnitude. As the angle values are continuous \change{quantity}{quantities}, we \add{have} quantized \change{the angle values}{those values}. For quantization, we tuned the number of quantization bins \add{as a} parameter. With experiments using various bin sizes (9, 16, 32 etc), we have found that using 16 bins with bin size 22.5 degree gives excellent results. After quantization to 16 bins, we compute co-occurrence matrix which is a $16\times 16$ matrix. We convert this $16\times 16$ matrix to a vector of size 256. This is our CoMOGrad feature vector. With this feature, we can simply take Euclidean distance to compare structures rather than using the alignment technique of $\alpha$ carbon distance matrices used by \textit{MASASW} and \textit{MatAlign}. Clearly, introduction of this feature makes the comparison method much simple\add{r} and faster.

\begin{figure}
 \centering
 \includegraphics[scale=0.9]{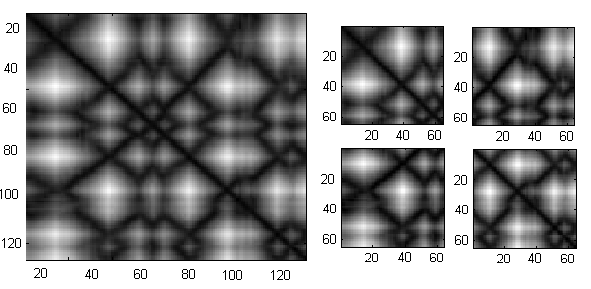}
  \caption{Level 1 quad tree of $\alpha$ carbon distance matrix image. \label{fig:PHOGEXTRACTION}}

\end{figure}

\paragraph{Pyramid histogram of oriented gradient (PHOG):}
\label{Phog}
The use of CoMOGrad gives us an ultra fast structure retrieval algorithm\change{,}{.} \change{it}{However it} achieves this speed at the cost of \add{some} reduction in accuracy. From the discussion in the previous sections, it is clear that we have to incorporate \add{the} gradient magnitude and spatial orientation of gradient along with \add{the} angular orientation of gradient to accurately describe the tertiary structure of \add{a} protein. The CoMOGrad feature only includes angular orientation of gradient and co-occurrence of angular orientation of gradient. To incorporate \add{the} gradient magnitude and spatial orientation of gradient along with angular orientation of gradient, we take another feature named pyramid histogram of oriented gradient (PHOG) together with our CoMOGrad feature to improve \add{the} accuracy. PHOG was first proposed by Bosch et al.~\cite{bosch2007representing} and successfully used in object classification and pattern recognition. We create a quad tree of the original image \add{with the original image at the root as follows.} \remove{, where we have the original image at the root, and} Each node \add{of the quad tree has} \remove{have} four children\add{, namely,} \remove{as} \emph{top-left} , \emph{top-right}, \emph{bottom-left} and \emph{bottom right}. Each of these images are of size one fourth of the original image. In Figure~\ref{fig:PHOGEXTRACTION}, we have shown \add{a} quad tree up to level 1. In our experiments, we have taken \add{the} quad tree up to level 3 and \change{found to have}{achieved} excellent results. For quad tree up to level 3, there are 1+4+4*4+4*4*4=85 nodes. We create gradient orientation histogram with 9 bins each with 40 degree range for each of the nodes. Now, we have 85*9=768 features. We incorporate these 768 features to a vector of size 768. Then, we normalize the vector by dividing it with the sum of its 768 components. This is our PHOG feature vector. Now, PHOG combined with CoMOGrad gives \add{a} total of 256+768=1021 features. \remove{As the distance measure of our new feature, we use the Euclidean distance or $\ell^2$ norm. The $\ell^2$ norm distance of the CoMOGrad and PHOG gives us much accurate and reasonably faster retrieval algorithm.}

\subsubsection*{Distance Measure}
We use the Euclidean distance or $\ell^2$ norm as the distance measure of our new features. PHOG combined with CoMOGrad can be seen as a vector of length 1021. Suppose, $f_q$ and $f_i$ denote the feature vectors of the query protein $q$ and a protein $i$ in the database, respectively.
Then the distance score $d_i^q$ of protein $q$ and $i$ would be calculated according to Eqn. \ref{eqn_dist} below.
\begin{equation}\label{eqn_dist}
d_i^q = \sqrt{\sum_{j=1}^{1021} ( |f_q[j]-f_i[j]|)^2}
\end{equation}
Clearly, the above distance measure can be calculated in $O(N)$ time where $N=1021$ is the size of our feature vector. Also, note that for CoMOGrad alone, $N = 256$ only. Our algorithm needs to compute $d_i^q$ for each protein $i$ in the database and then sort the results to rank them. As will be reported later, the $\ell^2$ norm distance of the CoMOGrad and PHOG gives us a fast and much accurate retrieval algorithm.

\section{Results and Discussion}
\label{experimental results}
Our method is readily available for use from this website: \url{http://research.buet.ac.bd:8080/Comograd/}. We have implemented our algorithm in Java (jdk 1.6) with Netbeans IDE and MySQL database. The feature extraction \change{was}{has been} done Using \textit{\change{Matlab}{MATLAB}} R2012. Our source code of our implementation is available from \url{https://github.com/rezaulnkarim/protein_tertiary_structure_retrieval-}. The experiments \change{was}{have been} run on  GNU/Linux debian ubuntu i686 operating system. \add{We have used a machine having} \remove{The machine configuration have been} \textit{Intel} (R) Core(TM) i5 3470 CPU 3.20 GHz with 4GB RAM. We have compared our methods with the tertiary structure retrieval method, \textit{MASASW}~\cite{6051424}\add{, which is shown to be the best performer in the literature to date}. We \add{have} implemented two versions of our method, one using  CoMOGrad and the other combining CoMOGrad with PHOG. 

\subsection{Benchmark Dataset\add{s}.}
For the experiments, we have used \textit{SCOP} \cite{scopURL,Murzin1995536} domains classification. SCOP is well accepted as the benchmark in the literature. We have taken 152,487 chains from \textit{SCOP} domain as the search space and extracted features from them. As the query proteins, we have used 4965 protein chains from 417 \textit{SCOP} family, 330 superfamily, 234 folds and 11 classes. We have \change{ran}{run} all 4965 queries over the search space of 152,487 \change{chains}{proteins} using our methods and sorted them all in ascending order based on \change{the}{our} distance measure. For MASASW \cite{6051424}, we ran the same experiment and sorted \change{them all}{results} in descending order based on \change{their}{its} similarity measure. Then \change{both were compared with}{we compare the results considering the top four} \textit{SCOP} labels, namely \emph{class, fold, superfamily} and \emph{family}. \concern{List of the query sequences are given in the Supplementary Table 1.}

\subsection{\change{Comparison of accuracy}{Accuracy Comparison}.}
\begin{figure}
 \begin{center}
 \includegraphics[scale=0.8]{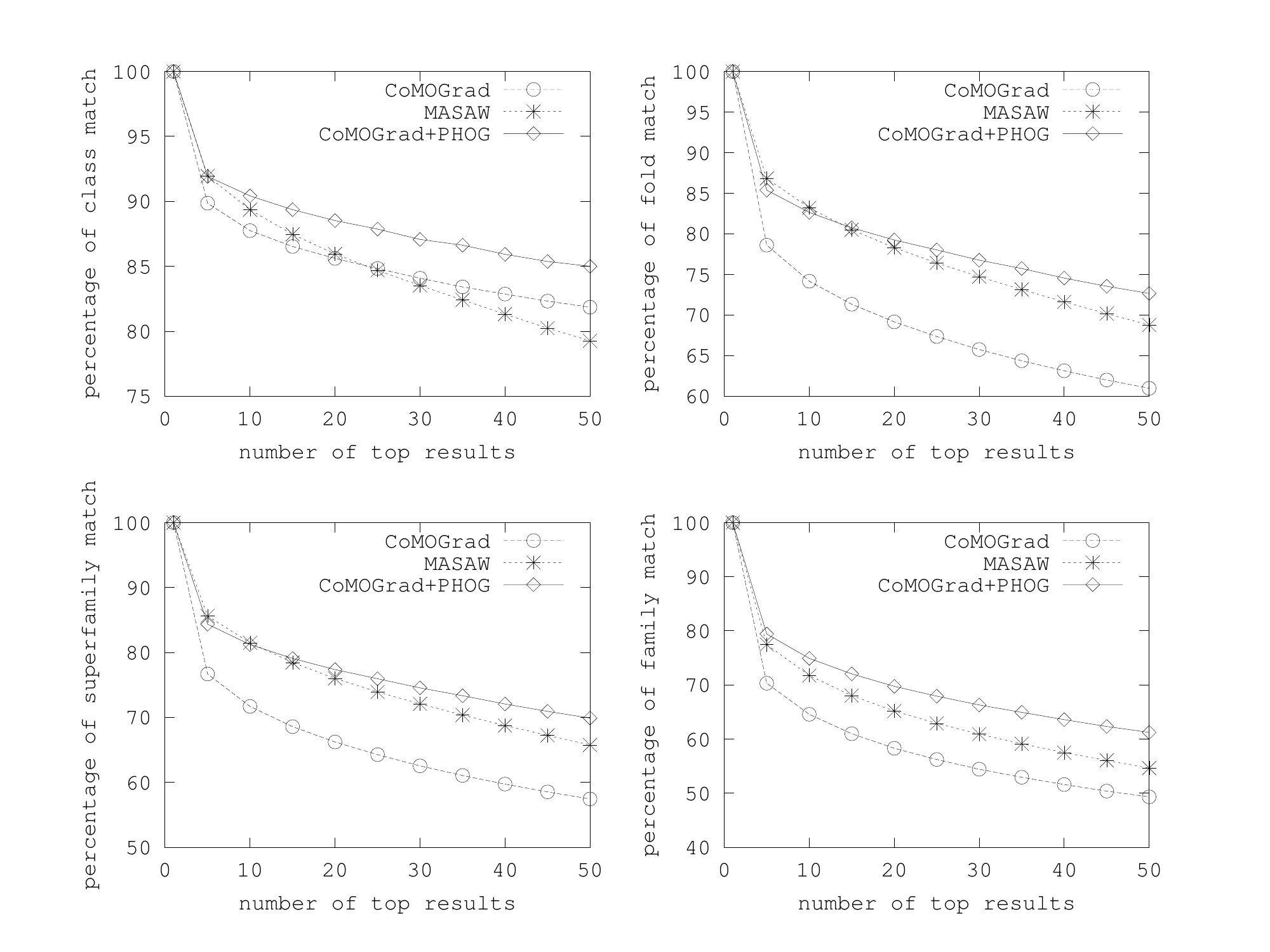}
 \end{center}
 \caption{Percentage of matches of Class, Fold, Superfamily and Family for up to top 50 retrieval results.\label{comogSuperFamilyAndFamily}}
 \end{figure}

The comparison of the accuracy of the query results \change{for}{among} \textit{MASASW} and our \add{two} methods are provided in Figure~\ref{comogSuperFamilyAndFamily}. In the line graphs plotted in \add{this} figure, the horizontal axis entitled ``\emph{number of top results}" \change{means}{tracks the} number of top ranked query results. The vertical axes in each figure \remove{is} entitled \remove{as percentage} ``\emph{\add{\%} of `label' match"} \remove{It} indicates the average number of query results that have matched the \textit{SCOP} `label' with \change{their}{the} corresponding query protein. \add{So, each point in the figure reports the average number query results that have matched the \textit{SCOP} `label' with the corresponding query protein for a specific number of top results.} We report the results for \add{all the \textit{SCOP} labels, i.e., \emph{class, fold, superfamily} and \emph{family}.} \remove{percentage of class match, percentage of fold match, percentage of superfamily match and percentage of family. They represent the matching of the top ranked results found with the \textit{SCOP} labels of the query protein. However, accuracy of the retrieval is defined as in the following equation:}


As an \change{explanation}{example}, suppose, we consider \add{the} top 50 results of the query and \add{have} run \add{a} total \add{of} 3 \change{query}{queries}. \add{Assume that,} among the query results, \add{the} numbers of \add{match for the label \emph{family} are} \remove{results whose family matched with that of query protein are viz} 40, 42\remove{,} \add{and} 48 \add{respectively}. Then \add{the} average number of family match is (40+42+48)/3=45 and \add{the} percentage of family match for \add{the} top 50 retrieval results is ($45 \times 100$ ) /(50) = $90$ percent. The line graphs are drawn with percentage of matches for class, fold, superfamily and family for \add{the} top 5 to top 50 retrieval results \change{for}{based on} the selected 4965 queries.

\subsection{\add{Runtime} Comparison \remove{of runtime}.}
The \change{required runtime for query}{time required} to retrieve \add{the} protein tertiary structure\add{s} \remove{with $\alpha$ carbon distance matrix feature using} \add{for all three} \remove{\textit{MASASW} algorithm and our} methods are \change{provided}{reported in} \remove{the} Table~\ref{tab:runtime comparision}. The run time is recorded \change{for}{by executing} 100 queries for each of the methods. \change{Here}{In the table,} \emph{Loading Time} is the time needed to load the feature vectors from disk to memory which is done once when the system starts. \add{The} \change{Average query time}{\emph{Time Per Query} in the table} is the time needed to compare the feature vector of a query protein with that of all the 152,487 proteins in our \change{system}{protein database} and sort the results based on \add{the} distance/similarity \add{measure} and \add{to} return the sorted \add{top} ranked results. \add{Note that the} \change{Average query time}{query time reported} excludes the loading time. The \emph{Query Time} in the table indicates the total time needed for 100 query structures. The results indicate\remove{s} that \change{average}{the} query time for the variant with only the CoMOGrad feature is ultra \change{first}{fast albeit at the cost of some reduction in the accuracy.} \remove{although it costs a reduction in accuracy.} However, the variant using both CoMOGrad and PHOG as features is both super \change{speedy}{fast} and more accurate than \textit{MASASW}. \remove{However, if the number of queries is increased, the ratio of run time for both of the variants of our algorithm compared to \textit{MASASW} will be increased more as for small number of queries the time for sorting the results which is same for all the methods have dominance.}
\begin{table}
\caption{Comparison of query time.}
\centering
    \begin{tabular}
    { | l  | c |c | c |}
    \hline
    Method   & Loading Time & Query Time & Time Per Query \\ \hline
    MASASW  & 28 min 11 s & 42 min 18 s   & 25.38 s . \\ \hline
    CoMOGrad  & 18 m 31s & 1 m 23 & 0.83 s \\ \hline
    CoMOGrad + PHOG  & 27 min 24 s & 5min 32 s & 3.32 s. \\ \hline
    \end{tabular}
\label{tab:runtime comparision}
    \end{table}
\subsection{Discussion.}
\label{performing_eval}
The exact \change{solution}{algorithm} for the matching of tertiary structures with $\alpha$ carbon distance matrix runs in $O(N!)$ time assuming that the input matrix is of size $N\times N$. The time complexity of \textit{MASASW} \add{for comparing two features} is \remove{of} $O(wWN^2)$\add{, which assumes } \remove{and they take} $N=32$ as the dimension of the distance matrix. \add{Here,} $W$ and $w$ are the size of \add{the} sliding windows to align matrices and to align rows, respectively. \change{They}{The authors of MASASW have} empirically obtained the most reasonable and suitable values for $W$ and $w$ which are $5$ and $8$ respectively. Our CoMOGrad feature is a vector of size $256$ and the combination of CoMOGrad and PHOG \change{uses}{gives us} a feature vector of size of $1021$.
\remove{From the results provided above, it is clear that $\ell^2$ norm distance or Euclidean distance of our feature vectors is much better than aligning the $32\times 32$ distance matrices.}
For both of \change{the}{our} methods, \remove{when these features are stored in the database,} the run time \add{for comparing two features} is just \change{$O(N^2)$}{$O(N)$}. Therefore, when only \add{the} CoMOGrad feature is used, the \change{runtime of our method}{time to compare two features} is approximately $(32 \times 32 \times 5 \times 8)/(256)$= $160$ times faster than \textit{MASASW}. \change{Also note that,}{And,} the combination of CoMOGrad and PHOG is approximately $(32 \times 32 \times 5 \times 8)/(1021)$= $40$ times faster than \textit{MASASW} \add{in this respect}. The feature extraction of \add{the} query protein \add{as and} when \change{query}{it} is submitted as \change{\emph{pdb} format}{a} coordinate file \add{in the \emph{PDB} format} does not have noticeable effect on \add{the} \change{run time}{running time} as this operation is done for one\add{, i.e., the} query protein \add{only;} \change{and}{the} features of all \change{protein}{the proteins} in \add{the} target search space \add{(i.e., in the database)} are {made available beforehand as they have already been} preprocessed. \remove{We have shown the exact query response time for all of the methods that further establish our claims.}

Compared to \textit{MASASW}, the query time of the variant with CoMOGrad is  almost 30 times and the variant with CoMOGrad and PHOG is more than 7 times faster. As seen in our earlier discussion, theoretically the variant with CoMOGrad is 160 times faster and the variant with CoMOGrad and PHOG is 40 times faster in comparing two features. However, in addition to the feature comparison, the actual retrieval algorithm needs to perform a sorting operation on the results of the distance/similarity values of all the proteins in the database against the query protein. Note that both MASASW and our two methods need to use this sorting algorithm. MASASW however sorts in descending order because it uses a similarity measure. As a result the actual improvement in the query time achieved by our methods does not completely match with the theoretical deduction. \remove{Clearly, if the number of queries is increased, the improvement in speed would be more prominent.}

In terms of accuracy, the performance of our method using \add{only} the CoMOGrad feature \remove{results} is almost \change{near}{similar} to that of \textit{MASASW}. Nonetheless, compared to \textit{MASASW}, the combination of CoMOGrad and PHOG \add{achieves} \remove{have} a higher \add{degree of} accuracy. The experimental dataset contains of proteins taken from 417 various families and \add{the} number is significant. The most significant achievement of the novel features we have used is the substantial reduction of the query processing time. Together the accuracy and reduced query processing time makes the structural comparison the simplest state of the art technique.

\section*{Conclusion}
\label{conclusion}
In this paper, we have presented two novel features\add{, namely,} CoMOGrad and PHOG\add{,} for faster \add{and accurate} retrieval of \add{the} protein tertiary structures. We have compared our results \change{with different}{considering all the} levels of \textit{SCOP} classification hierarchy. We have \change{showed}{reported} average percentage of matching for class, fold, super family and family of our retrieval results with the query protein while most of the works \add{in the literature have only shown} \remove{showed} similarity on only class and fold\add{;} \remove{and} very few \add{in fact have} worked \change{for}{on} automated similarity match for the lower levels. Our results are in good compliance with \add{the} \textit{SCOP} classification.  CoMOGrad feature is ultra fast \add{as} compared to \add{the} state of the art methods but \change{it costs}{this extreme speed is achieved at the cost of a} slight reduction in \add{the} accuracy. The combination of CoMOGrad with PHOG is \change{super}{also very} fast and \change{even}{at the same time is} superior in \add{terms of} accuracy \add{as} compared to state of the \add{art} methods. This creates the possibility to implement \change{an online}{a} web based service for \add{the} protein tertiary structure retrieval \add{with a truly online behaviour, i.e., that can provide the results in seconds,} while the present web services \add{usually} provide\remove{s} query results via email only.


\bibliographystyle{abbrv}
\bibliography{nature2014}


\section*{Acknowledgements}
The authors thanks Georgina Mirceva for sharing the implementation details of the MASASW algorithm.


\end{document}